\pdfoutput=1
\pdfoutput=1
\pdfoutput=1
\pdfoutput=1
\pdfoutput=1
\documentclass[journal, comsoc]{IEEEtran}
\usepackage[T1]{fontenc}
\usepackage{amsfonts,amsmath,amssymb,graphicx,dsfont,color,multirow,epstopdf}
\usepackage{amsmath,bm}
\usepackage{color}
\usepackage{cite}
\usepackage{hyperref}
\usepackage{cleveref}

\ifCLASSINFOpdf
\else
\fi
\usepackage[square, comma, sort&compress, numbers]{natbib}
\usepackage{amsmath}
\usepackage[cmintegrals]{newtxmath}
\begin{document}

\title{Mining Generalized Features for Detecting AI-Manipulated Fake Faces\thanks{This work was supported in part by the National Key Research and Development of China (2018YFC0807306), the National Science Foundation of China (61672090, U1936212), and the Fundamental Research Funds for the Central Universities (2018JBZ001). }% <-this % stops a space
	\thanks{Corresponding author: Rongrong Ni }
	\thanks{Yang Yu, Rongrong Ni and Yao Zhao are with the Institute of Information Science, Beijing Jiaotong University, Beijing 100044, China, and also with the Beijing Key Laboratory of Advanced Information Science and Network Technology, Beijing 100044, China (e-mail: 18112012@bjtu.edu.cn; rrni@bjtu.edu.cn; yzhao@bjtu.edu.cn).}}% <-this % stops a space
\author{Yang Yu, Rongrong Ni and Yao Zhao,~\IEEEmembership{Senior Member, IEEE}}% <-this % stops a space
\maketitle
\IEEEpeerreviewmaketitle

\begin{abstract}

Recently, AI-manipulated face techniques have developed rapidly and constantly, which has raised new security issues in society. Although existing detection methods consider different categories of fake faces, the performance on detecting the fake faces with "unseen" manipulation techniques is still poor due to the distribution bias among cross-manipulation techniques. To solve this problem, we propose a novel framework that focuses on mining intrinsic features and further eliminating the distribution bias to improve the generalization ability. Firstly, we focus on mining the intrinsic clues in the channel difference image (CDI) and spectrum image (SI) from the camera imaging process and the indispensable step in AI manipulation process. Then, we introduce the Octave Convolution (OctConv) and an attention-based fusion module to effectively and adaptively mine intrinsic features from CDI and SI. Finally, we design an alignment module to eliminate the bias of manipulation techniques to obtain a more generalized detection framework. We evaluate the proposed framework on four categories of fake faces datasets with the most popular and state-of-the-art manipulation techniques, and achieve very competitive performances. To further verify the generalization ability of the proposed framework, we conduct experiments on cross-manipulation techniques, and the results show the advantages of our method.
\end{abstract}


% \fi
\IEEEpeerreviewmaketitle

\section{Introduction}
\begin{figure}[ht]
	\centering
	\includegraphics[scale=0.49]{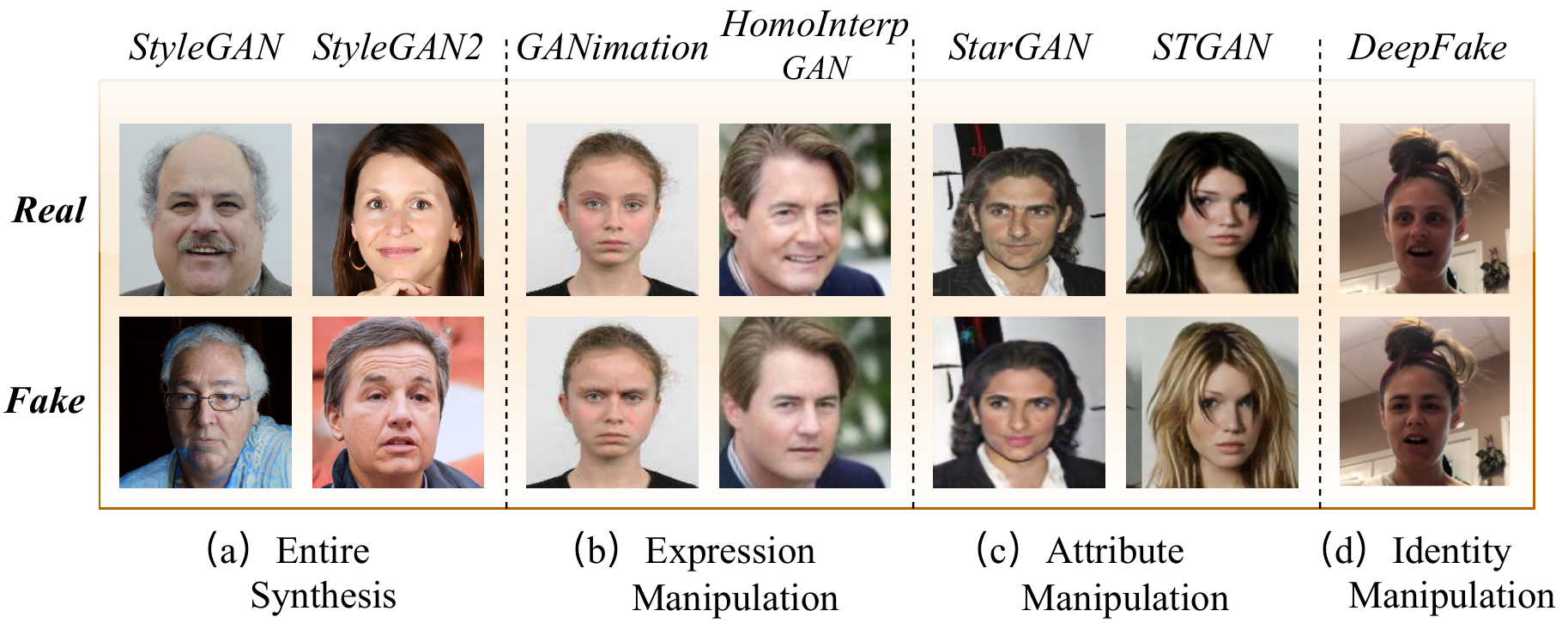}
	\caption{Four categories of fake faces (bottom) with various AI-manipulated techniques and real faces (top). (a) Entire Synthesis. (b) Expression Manipulation. (c) Attribute Manipulation. (d) Identity Manipulation.}
	\label{example}
\end{figure}

\IEEEPARstart{H}{uman} face images and videos contain personal information and play an important role in daily life, such as communication, access control and payment. However, with the remarkable development of AI-manipulated techniques, it is becoming increasingly easy to produce fake faces. Unlike previous simple face manipulation techniques (e.g., splicing), AI-manipulated techniques can easily produce more realistic fake face images, even fake face videos. Specifically, these techniques can synthesize non-existent face images \cite{karras2019style,karras2018progressive,karras2020analyzing} or directly manipulate face expressions images \cite{ding2018exprgan,pumarola2018ganimation,chen2019homomorphic, thies2016face2face}, face attributes images \cite{choi2018stargan,he2019attgan,liu2019stgan}, even manipulate identities in videos \cite{korshunova2017fast,petrov2020deepfacelab,thies2019deferred}. Fig.~\ref{example} presents four categories of fake faces with various AI-manipulated techniques, by which people can be easily fooled. These realistic fake faces may be abused for malicious purpose, raising security and privacy issues in our society. 
More critically, the detection methods of previous simple face manipulation techniques \cite{ chen2015automatic, dang2019face, zhao2014passive, liu2019adversarial} cannot work on the AI-manipulated fake faces. Therefore, it is extremely necessary to develop effective methods for detecting AI-manipulated fake faces.

\begin{figure*}[ht]
	\centering
	\includegraphics[scale=0.76]{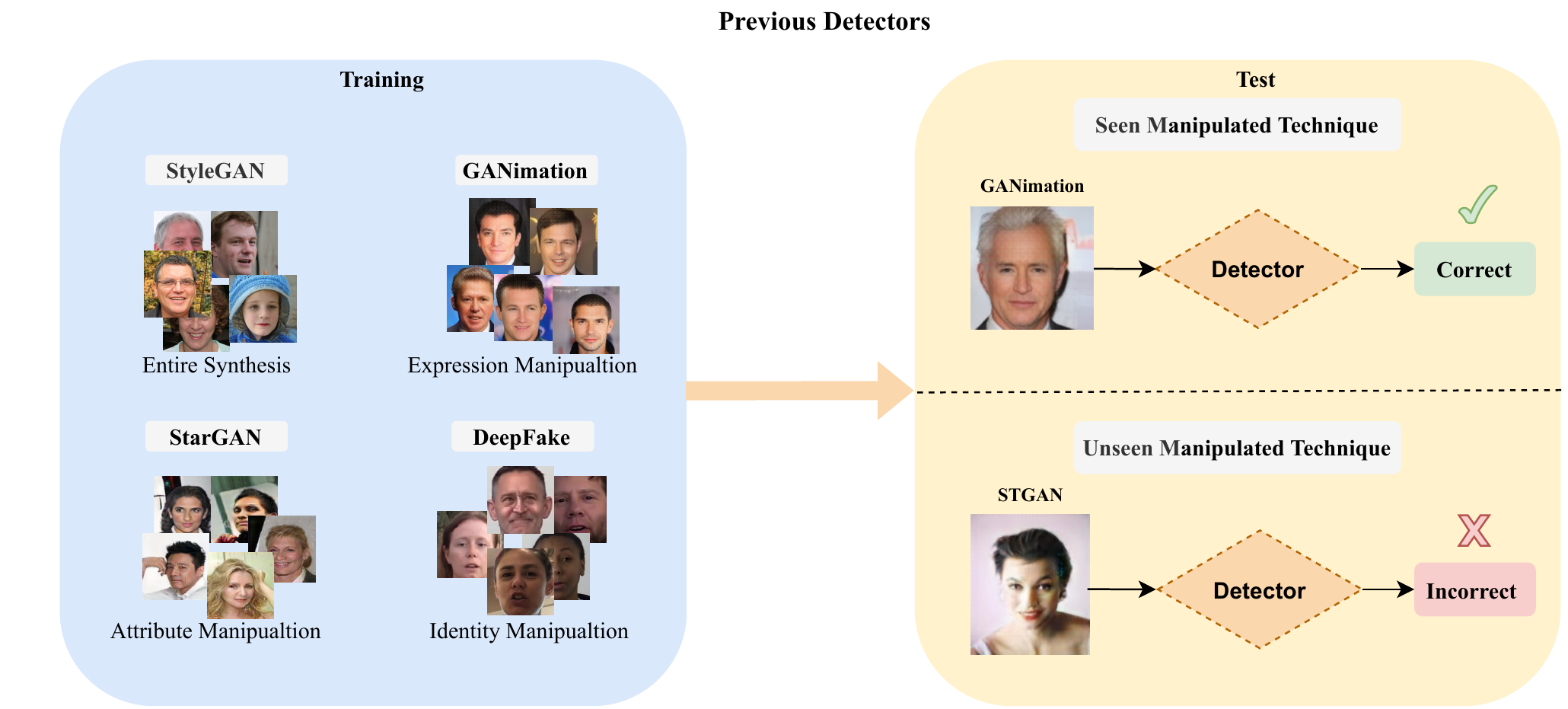}
	\caption{The previous detecors cannot perform well on the fake faces produced by unseen manipulation techniques.}
	\label{unseen}
\end{figure*}

The security concerns have motivated a number of studies for detecting AI-manipulated fake faces. Some handcrafted features-based detection methods heavily depend on the specific defects in the manipulation process and cannot extract intrinsic features (e.g., eye color variance \cite{matern2019exploiting} and lack of eye blinking \cite{li2018ictu}), which determines the short-term effectiveness of these methods. Some simple learning-based detection methods \cite{afchar2018mesonet,yu2019attributing,rossler2019faceforensics++,nguyen2019capsule,guera2018deepfake, sabir2019recurrent,amerini2019deepfake} aim to deal with certain categories of manipulation faces, but their effectiveness is limited to the certain categories they are trained for. To further consider the detection of all categories of fake faces, some learning-based methods focus on finding relatively common variances (e.g., neuron behaviors variance \cite{wang2019fakespotter} ) between real and fake faces and taking advantage of incremental learning \cite{marra2019incremental} or domain adaptation \cite{cozzolino2018forensictransfer} to detect fake faces continuously. Nevertheless, due to the distribution bias among cross-manipulation techniques, the above detection methods perform poorly when applied to the fake faces produced by "unseen" manipulation techniques, as shown in Fig.~\ref{unseen}. Thus, how to improve the generalization ability on detecting the fake faces with unseen manipulation techniques becomes increasingly challenging. To address this issue, several attempts have been proposed recently. Some approaches attempt to simulate and generalize to unseen fake face generators with one specific architecture (e.g., AutoGAN \cite{zhang2019detecting} and ProGAN \cite{wang2020cnn}). Some approaches attempt to utilize the attention mechanism to highlight the informative regions to reduce the interference of bias mentioned above \cite{dang2020detection}. Some approaches assume the existence of a blending step in the fake face manipulation process (e.g., Face X-ray \cite{li2020face}). However, since the intrinsic features cannot be captured well enough and the assumption is too strong, the above methods cannot perform well on the fake faces produced by new emerging manipulation techniques.

In this paper, we aim to improve the generalization ability on detecting fake faces with unseen manipulation techniques. As analyzed above, there are still two important issues on improving generalization ability. On one hand, the intrinsic features in all categories of fake faces produced by various manipulation techniques still need to be mined. On the other hand, how to eliminate the bias of manipulation techniques becomes increasingly pivotal. In view of this, we design a novel framework that focuses on mining the intrinsic features and further reducing the bias mentioned above to obtain a more generalized framework. Firstly, for fake faces with various AI-manipulated techniques, we mine two intrinsic clues in the channel difference image (CDI) and spectrum image (SI) view of the camera imaging process and the indispensable step in AI manipulation process, rather than depending on the specific defects in the manipulation process. In particular, we find both clues are related to frequency domain information, thus the Octave Convolution (OctConv) \cite{chen2019drop} which has been proved to be efficient for capturing frequency information is employed to learn intrinsic features for detecting AI-manipulated fake faces from CDI and SI. Moreover, an attention-based fusion module is exploited to adaptively weight features, guiding the effective performance of fused features. Finally, to obtain a more generalized framework, we design an alignment module to eliminate the distribution bias among cross-manipulation techniques. 
 
 Our main contributions can be summarized as:

$ \bullet  $ We mine intrinsic clues from the camera imaging process and the indispensable step in AI manipulation process and further adopt OctConv and an attention-based fusion module to effectively mine intrinsic features for detecting AI-manipulated fake faces.

$ \bullet  $ To further improve generalization ability for our framework, we design an alignment module to reduce the bias by minimizing the difference in feature distribution among cross-manipulation techniques.

$ \bullet  $ We conduct extensive evaluations on four categories of fake faces datasets as well as our proposed cross-manipulation technique based protocols to verify the comprehensiveness and generalization ability of our detection framework. The results show the effectiveness of our proposed framework compared with other state-of-the-art methods, especially on the generalization performance. Extensive ablation studies demonstrate the effectiveness of each component in our framework.

The rest of the paper is organized as follows. Related works are briefly reviewed in Section~\ref{related}. Analysis of intrinsic clues and the proposed framework for mining generalized features to detect AI-manipulated fake faces are presented in Section~\ref{method}. Section~\ref{experiment} shows the experimental results and corresponding analysis. The conclusion and future studies are drawn in Section~\ref{con}.

\section{Related Works}
\label{related}
In this section, we discuss the most relevant methods including fake face manipulation techniques and fake face detection methods.

\subsection{Fake Face Manipulation Techniques}

Existing AI-manipulated fake faces can be roughly categorized into four categories: entire face synthesis, facial expression manipulation, facial attribute manipulation, and face identity manipulation. For entire face synthesis, various Generative Adversarial Networks (GANs) are usually used to create entire non-existent faces. Karras \textit{et al}. proposed StyleGAN \cite{karras2019style} as an improved version of their previous popular approach ProGAN \cite{karras2018progressive}, which introduced an alternative generator architecture to synthesize highly varied and higher-quality human faces. Then they presented the StyleGAN2 \cite{karras2020analyzing} to further improve the quality of the generated face images. For facial expression manipulation, powerful GANs and 3D face reconstruction methods are widely used for this manipulation. Ding \textit{et al}. \cite{ding2018exprgan} proposed ExprGAN for photo-realistic facial expression manipulation with controllable expression intensity. Albert \textit{et al}. \cite{pumarola2018ganimation} introduced a GAN conditioning scheme GANimation based on Action Units (AU) annotations, which manipulated human expression in a continuous manifold. Chen \textit{et al}. \cite{chen2019homomorphic} proposed HomoInterpGAN to generate high-quality results for unpaired facial expression translation. Thies \textit{et al}. \cite{thies2016face2face} presented Face2Face for real-time facial expressions reenactment from one person to another via re-render and animation methods. For facial attribute manipulation, this manipulation edits single or multiple attributes in a face (e.g., gender, age, skin color, hair, and glasses), usually through GAN-based frameworks for general image translations and manipulations. Choi \textit{et al}. \cite{choi2018stargan} proposed StarGAN to perform image-to-image translations for multiple domains using a conditional attribute transfer network and achieved good visual results. He \textit{et al}. \cite{he2019attgan} proposed AttGAN which provided realistic attribute manipulation results with other facial details well preserved by applying the attribute classification constraints. Liu \textit{et al}. \cite{liu2019stgan} proposed STGAN to improve the attribute manipulation ability and the image quality by incorporating selective transfer units with encoder-decoder. FaceApp\footnotemark[1] popularized facial attribute manipulation as a consumer-level application, which provided 28 filters to modify specific attributes. For face identity manipulation, this manipulation replaces the face of one person with the face of another person. Two different approaches are usually considered: classical computer graphics-based techniques such as FaceSwap \cite{korshunova2017fast}, and novel deep learning techniques known as DeepFakes \cite{petrov2020deepfacelab} and NeuralTextures \cite{thies2019deferred}.

\footnotetext[1]{https://apps.apple.com/gb/app/faceapp-ai-face-editor/id1180884341}

\subsection{Fake Face Detection Methods}
Fake face manipulation detection methods can be broadly classified into two classes, i.e., handcrafted features-based methods and deep learning-based methods.

The handcrafted features-based methods try to highlight specific defects in the fake face manipulation process. Matern \textit{et al}. \cite{matern2019exploiting} detected DeepFakes and Face2Face videos based on visual artifacts, such as eye color variance, unconvincing specular reflections and missing details in the eye and teeth areas. Li \textit{et al}. \cite{li2018ictu} proposed to expose AI-created fake videos by detecting the lack of eye blinking of the DeepFake videos. Then, they also utilized face warping artifacts \cite{li2019exposing}, face landmark locations \cite{yang2019exposing}, and inconsistent head pose \cite{yang2019inconsistent} to expose DeepFake videos. Fernandes \textit{et al}. \cite{fernandes2019predicting} revealed DeepFake based on the lack of variations induced by heart beating. McCloskey \textit{et al}. \cite{mccloskey2019detecting} analyzed that the GAN-generated images lack saturated regions. These detection methods rely on the handcrafted features which mostly depend on the specific defects in the manipulation process, thus the main drawback of these approaches remains that they are likely to soon become ineffective as generation methods evolve. Therefore, the intrinsic features of AI-manipulated fake faces still need to be mined.

For the learning-based methods, some simple CNN-based methods were proposed primarily. Afchar \textit{et al}. \cite{afchar2018mesonet} detected DeepFakes and Face2Face videos via two networks (Meso-4 and MesoInception-4) with a low number of layers that focus on the mesoscopic properties of images. Yu \textit{et al}. \cite{yu2019attributing} presented a method to detect fake images by learning GAN artificial fingerprints. Rossler \textit{et al}. \cite{rossler2019faceforensics++} introduced a face manipulation dataset FaceForensics++ and utilized Xception \cite{chollet2017xception} to improve forgery detection accuracy in the presence of strong compression. Nguyen \textit{et al}. \cite{nguyen2019capsule} introduced a capsule network to detect forged images and videos. The recurrent neural network \cite{guera2018deepfake, sabir2019recurrent} and the optical flow \cite{amerini2019deepfake} were adopted to utilize the time information for exposing fake face videos. Mi \textit{et al}. \cite{mi2020gan} equipped the algorithm with a much better comprehension of the global information with the self-attention mechanism to detect entire fake images. Liu \textit{et al}. \cite{liu2020global} presented Gram-Net that leveraged global image texture representations for generalization ability promotion on detecting the fake faces. He \textit{et al}. \cite{he2019detection} employed the ensemble of deep representations from multi color spaces for detecting fake images and further applied the random forest classifier against different post-processing attacks. To further consider the detection ability on all categories of fake faces, Wang \textit{et al}. \cite{wang2019fakespotter} introduced FakeSpotter to spot AI-manipulated fake faces by monitoring neuron behaviors. In order to continuously detect fake faces, Marra \textit{et al}. \cite{marra2019incremental} proposed an incremental learning detection method and Cozzolino \textit{et al}. \cite{cozzolino2018forensictransfer} introduced an autoencoder-based ForensicTransfer to detect fake face with novel manipulation techniques using a few examples, without worsening the performance on the previous ones. Qian \textit{et al}. \cite{qian2020thinking} proposed $ F^{3} $-Net to detect fake faces based on  two different but complementary frequency-aware clues. However, the effectiveness of these methods was limited to the manipulation techniques they were trained for, and most of these detectors perform poorly on the unseen manipulation techniques.

%methods were used to detect fake face videos in the optical flow \cite{guera2018deepfake, sabir2019recurrent}. Amerini \textit{et al}. \cite{amerini2019deepfake} adopted the optical flow fields to exploit possible inter-frame dissimilarities for exposing fake face videos.
\begin{figure*}[ht]
	\centering
	\includegraphics[scale=0.175]{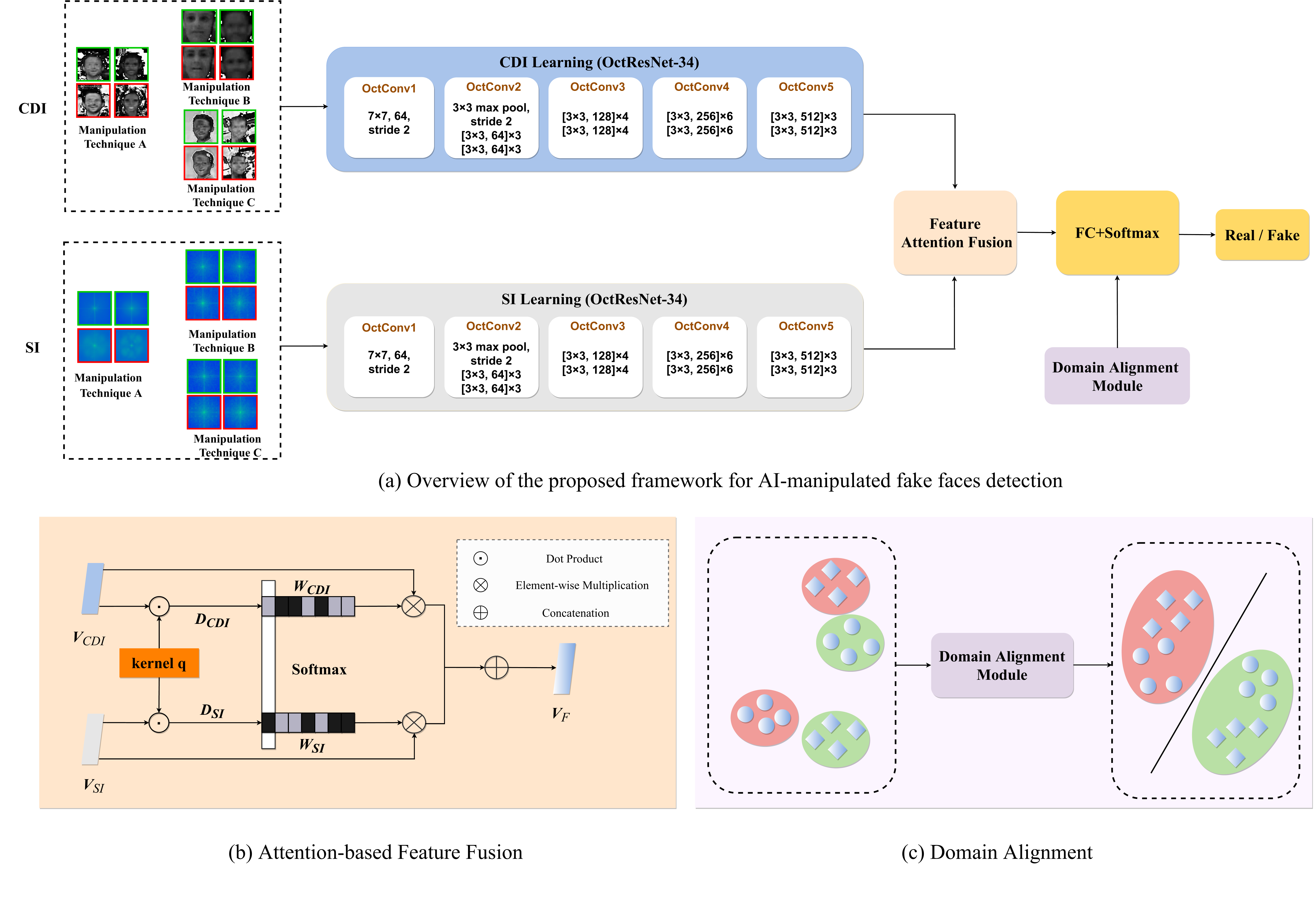}
	\caption{Overview of the proposed framework for mining generalized features to detect AI-manipulated fake faces. This framework consists of three phases: First, the feature learning module accepts CDI and SI as inputs to mine intrinsic features with OctResNet-34. Then, the feature attention fusion module is adopted to adaptively fuse these two features. Finally, the domain alignment module is specially proposed to reduce the bias of manipulation techniques to by minimizing the difference in feature distribution among cross-manipulation techniques, thereby further improving the generalization ability of our framework on detecting fake faces with unseen manipulation techniques.}
	\label{framework}
\end{figure*}

\begin{figure*}[ht]
	\centering
	\includegraphics[scale=0.8]{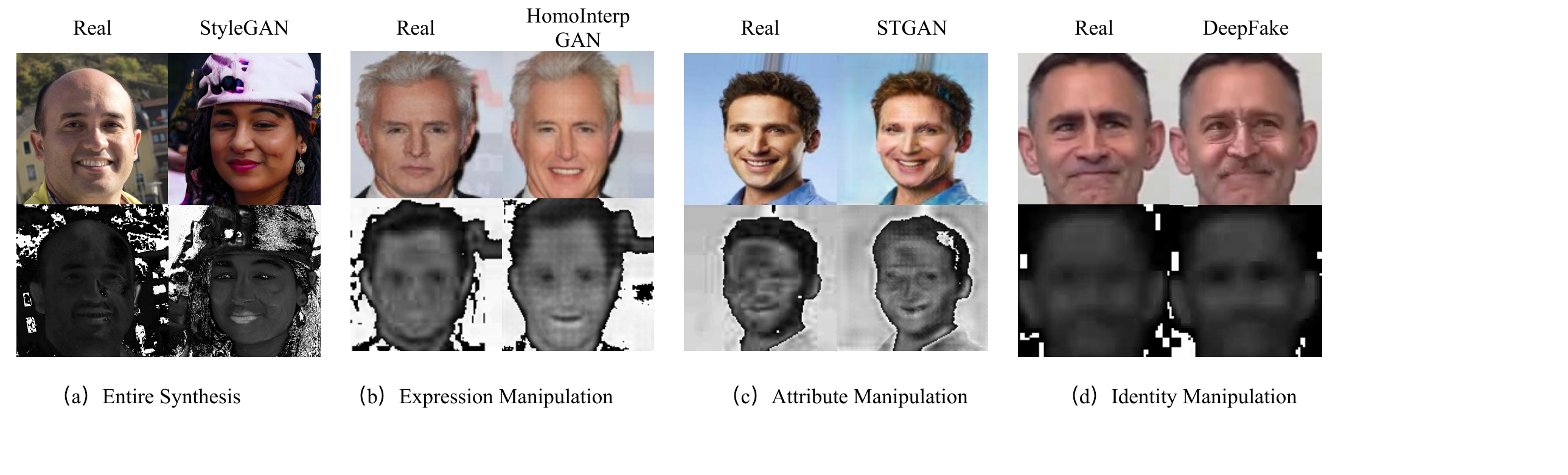}
	\caption{ The real face images and the corresponding $ R-G $ CDIs (left), and four categories of fake faces with different manipulation techniques and the corresponding $ R-G $ CDIs (right). (a) Entire Synthesis. (b) Expression Manipulation. (c) Attribute Manipulation. (d) Identity Manipulation.}
	\label{CDI}
\end{figure*}

\begin{figure*}[ht]
	\centering
	\includegraphics[scale=0.8]{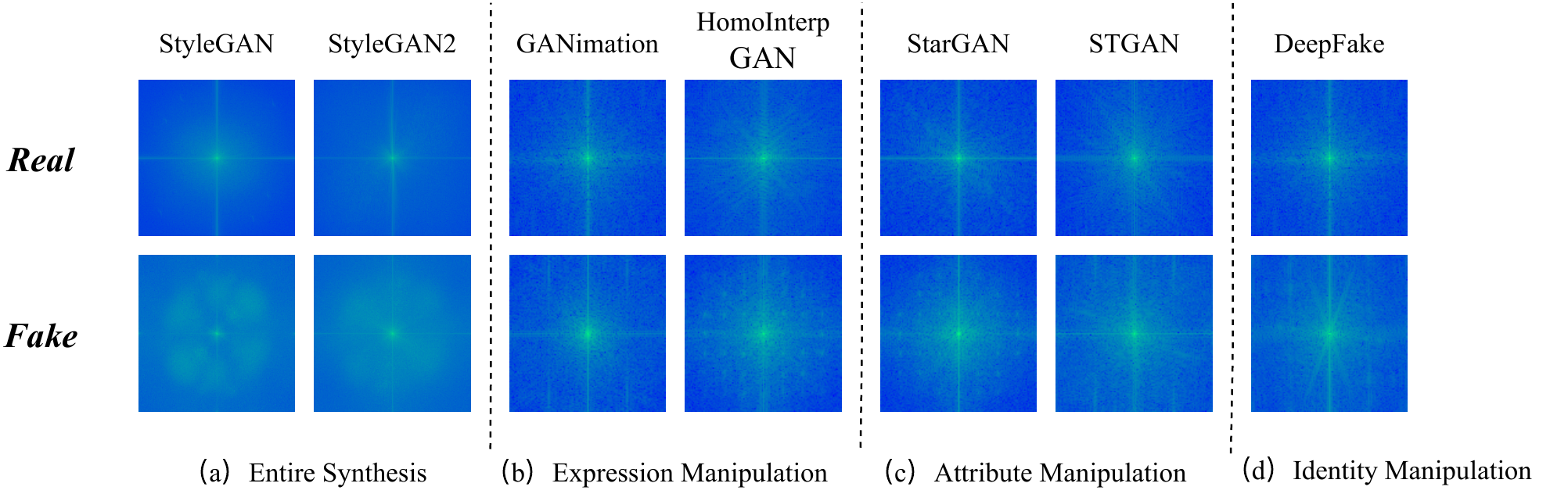}
	\caption{Average spectrum images (SI) of four categories of fake faces (bottom) with different manipulation techniques and corresponding real faces (top). (a) Entire Synthesis. (b) Expression Manipulation. (c) Attribute Manipulation. (d) Identity Manipulation.}
	\label{SI}
\end{figure*}

Some approaches focus on detecting the fake faces with unseen manipulation techniques. Xuan \textit{et al}. \cite{xuan2019generalization} proposed a pre-processing step to reduce low level artifacts of GAN images and force the discriminator to learn more general forensic features for improving the generalization ability. Li \textit{et al}. \cite{li2020face} found that most existing face manipulation techniques shared a common blending operation, thus proposed the Face x-ray to focus on the boundary of forged faces instead of the type of manipulation. Zhang \textit{et al}. \cite{zhang2019detecting} proposed a GAN simulator to reproduce common GAN-image artifacts, which manifested as spectral peaks in the Fourier domain. Li \textit{et al}. \cite{li2020identification} observed that the fake images were more distinguishable from real ones in the chrominance components, especially in the residual domain. Dang \textit{et al}. \cite{dang2020detection} utilized an attention mechanism to process and improve the feature maps for the fake face manipulation detection task. Wang \textit{et al}. \cite{wang2020cnn} detected GAN-generated images using a universal detector with careful pre and post-processing and data augmentation. Chai \textit{et al}. \cite{chai2020makes} used a patch-based classifier with limited receptive fields to visualize which regions of fake images were more easily detectable and further showed a technique to exaggerate these detectable properties. However, these methods cannot work well on the fake faces with the state-of-the-art manipulation methods.

%Fernando \textit{et al}. \cite{fernando2019exploiting} proposed a Hierarchical Memory Network by utilising knowledge stored in neural memories as well as visual cues to reason about the perceived face and anticipate its future semantic embeddings to render a generalisable face tampering detection framework.In \cite{du2019towards}, Locality-aware AutoEncoder with a pixel-wise mask was used to  learn intrinsic representation from the forgery region, so as to improve the generalization ability. Chen \textit{et al}. \cite{chen2020manipulated} proposed manipulated face detector based on features both in spatial domain and frequency domain and further added attention-based layers to improve its generalization ability. 

\section{Proposed Method}
\label{method}
In this section, we illustrate the proposed framework for mining generalized features to detect AI-manipulated fake faces in detail. The pipeline of this framework is shown in Fig.~\ref{framework}. Firstly, we mine the two intrinsic clues from the CDI and SI. For more details and the theoretical analysis, refer to the following Section~\ref{Analysis}. In Section~\ref{CDSI}, we introduce the features learning module that perceives these two sources of information to mine intrinsic features. Subsequently, the attention-based feature fusion module that adaptively fuse features is described. In Section~\ref{generalization}, to further obtain a generalized framework, we specially propose an alignment module to eliminate the bias among cross-manipulation techniques in feature distribution.

\subsection{Intrinsic Clues Analysis}
\label{Analysis}
\subsubsection{CDI Analysis }
In this part, we focus on mining the clues from the camera imaging process. In natural images, the high-frequency components across different color channels are highly mutually correlated and approximately equal due to the CFA interpolation algorithm during imaging process \cite{gunturk2002color}. Hence, we conduct the following analysis: one color channel can be formally described as:
\begin{equation}
I_{c}=I_{c}^{l}+I_{c}^{h}
\end{equation}
where $ c\in \left \{R,G,B\right \} $, and $ h $ and $ l $ denote the high-frequency and low-frequency components of image color channels. 

The channel difference can be expressed as:
\begin{equation}
I_{c_{1}}-I_{c_{2}}=I_{c_{1}}^{l}+I_{c_{1}}^{h}-I_{c_{2}}^{l}-I_{c_{2}}^{h}
\end{equation}
where $ c_{1 }$ and $ c_{2} $ represent different color channels.

For real faces, due to the the similarity of high-frequency components, $ I_{c_{1}}^{h}\approx I_{c_{2}}^{h} $. Therefore, the channel difference can be denoted as:
\begin{equation}
I_{c_{1}}-I_{c_{2}}\approx I_{c_{1}}^{l}-I_{c_{2}}^{l}
\end{equation}
As we can observe from Eq. (3), in the channel difference of real faces, the corresponding high-frequency components are filtered out, and only the low-frequency components are retained.

For entire-synthesis fake faces, there is no CFA interpolation algorithm in the generation process, which is different from the camera imaging process, thus the correlation of high-frequency components in color channels does not exist. For other categories of fake faces, because the change of values in three channels are different after the manipulation operation, thus the correlation of high-frequency components in color channels is destroyed. In both cases, the channel difference can not be expressed as Eq. (3). Therefore, the channel difference of four categories of fake faces contains more high-frequency components compared to real faces. Four examples are shown in Fig.~\ref{CDI} to verify this analysis, we calculate the channel difference images (CDI) of $ R-G $ of real faces and four categories of fake faces, respectively. In all types of fake faces, the CDIs of $R-G$ contain more face details than the real ones. Therefore, the CDI contains discriminative intrinsic information for detecting fake faces.

%\begin{equation}
%F(u, v)=\frac{1}{M N} \sum_{x=0}^{M-1} \sum_{y=0}^{N-1} f(x, y) e^{-j 2 \pi\left(\frac{u x}{M}+\frac{v y}{N}\right)}
%\end{equation}

\subsubsection{SI Analysis}
In this part, we focus on mining clues from the indispensable step in the process of generating AI fake faces. We study the pipeline of producing four categories of fake faces and find out the up-sampling modules are consistent. Zhang \textit{et al}. \cite{zhang2019detecting} show that the up-sampling results in replications of spectra in the frequency domain, thus we mine intrinsic clues from the frequency spectrum. We check the average frequency spectrum obtained by Discrete Fourier Transform (DFT) from four types of fake faces and corresponding real faces to study the artifacts. For each manipulation technique, we choose 2000 face images/frames randomly. Compared to the real faces, we find that there are obviously grid-like patterns in the frequency spectrum of fake faces generated by different manipulation techniques, as shown in Fig.~\ref{SI}. The reason is that up-sampling actually replicate multiple copies in the spectrum of low-resolution images over high-frequency parts in the spectrum of final high-resolution images. In addition, the DeepFake faces lose high-frequency information compared to genuine ones due to interference of more pre- and post-processing in the video forgery process, hence the spectrum contains less high-frequency components, as shown in Fig.~\ref{SI} (d). Therefore, the spectrum image (SI) also contains discriminative intrinsic clues for detecting fake faces.

%The reason is that up-sampling results in replications of spectra in the frequency domain, a clue considered by \cite{zhang2019detecting}. 

%Specifically, for an image, we first apply the 2D DFT to each RGB channel and get 3 channels of frequency spectrum. Then, we shift the spectrum so that the low frequency components are at the center of the spectrum. Finally, we compute the logarithmic spectrum and normalize the logarithmic spectrum to $ [-1, 1] $. 

%The frequency spectrum show the intensity of the pixel value change in the image, and it is helpful to analyze the high frequency and low frequency components of the image. 

According to above analysis, the discriminative clues in the CDI and SI rely on the fundamental differences between the camera imaging process and the AI-manipulated process, instead of depending on the specific defects in the manipulation process. Consequently, the CDI and SI could be used for mining intrinsic features for improving the generalization ability on detecting AI-manipulated fake faces.

\begin{figure}[ht]
	\centering
	\includegraphics[scale=0.53]{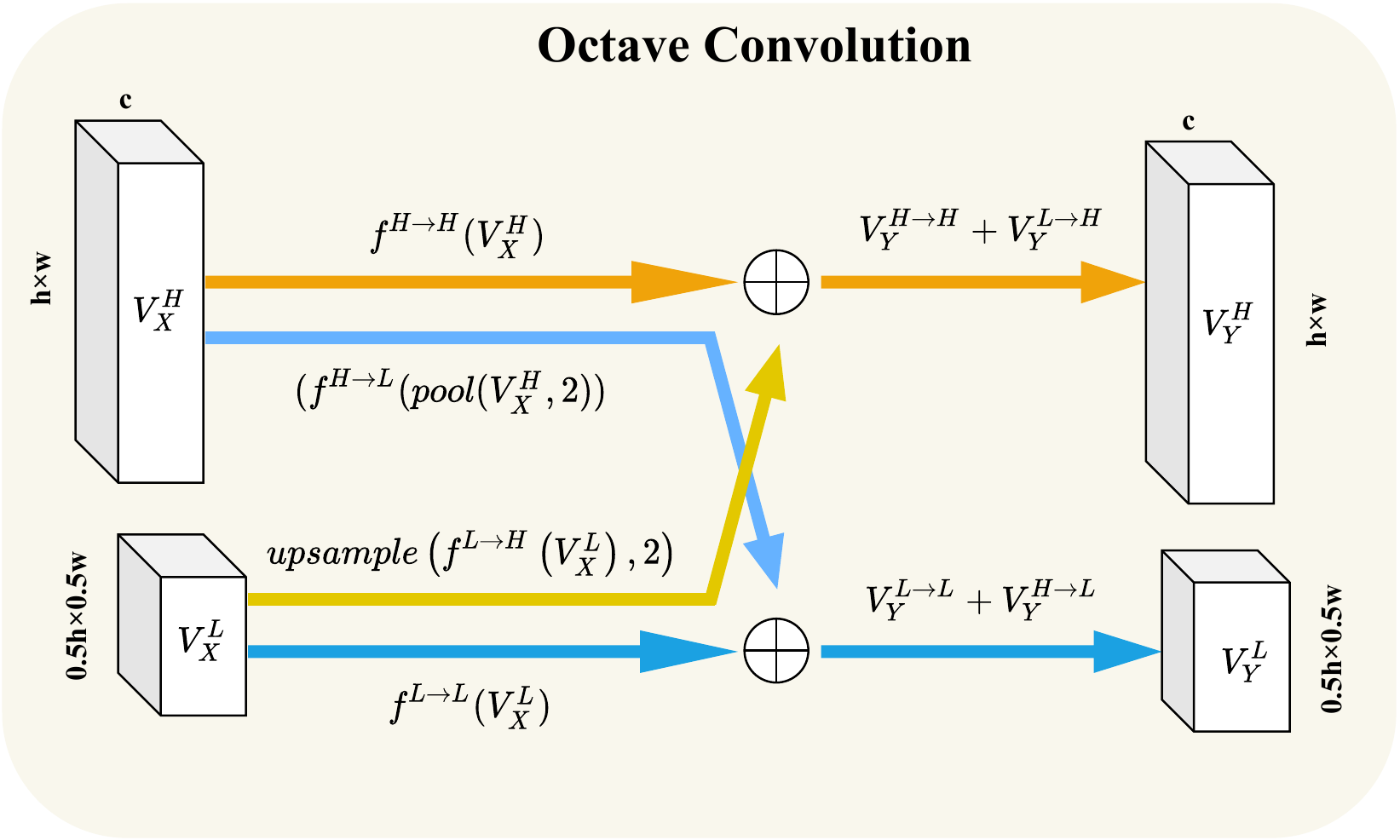}
	\caption{The details of the OctConv operator implementation.}
	\label{Oct}
\end{figure}

\subsection{ Intrinsic Features Mining }
\label{CDSI}
As mentioned above, both the CDI and SI contain discriminative intrinsic information for detecting AI-manipulated fake faces. Therefore, rather than roughly feeding RGB images into framework, we introduce the OctConv-based feature learning module and the attention-based fusion module to effectively mine intrinsic features from CDI and SI.

\subsubsection{Intrinsic Features Learning}
In the intrinsic features learning phase, we first construct CDI and SI of face images as two sources, as shown in Fig.~\ref{framework} (a), and note that we calculate the CDI of $R-G$, $B-G$, and $R-B$ for one face image and stack them in the channel dimension for dimensional uniformity. As analyzed above, both two clues are related the frequency domain information, thus we specially adopt the identical OctResNet-34 (ResNet-34 \cite{He2016Deep} with OctConv) to learn features from CDI and SI. The OctConv separately processes high- and low-frequency feature maps of the input $\textbf{\textit{V}}_{X}=\left\{\textbf{\textit{V}}_{X}^{H}, \textbf{\textit{V}}_{X}^{L}\right\}$, and further reduces the spatial resolution of low-frequency feature maps by an octave. The details of the OctConv operator implementation are illustrated in Fig.~\ref{Oct}. The high- and low-frequency feature maps of the output $\textbf{\textit{V}}_{Y}=\left\{\textbf{\textit{V}}_{Y}^{H}, \textbf{\textit{V}}_{Y}^{L}\right\}$ in OctConv can be described as follows:

\begin{equation}
\begin{aligned}
\textbf{\textit{V}}_{Y}^{H}=& \textbf{\textit{V}}_{Y}^{H \rightarrow H}+\textbf{\textit{V}}_{Y}^{L \rightarrow H} \\
=& f^{H \rightarrow H}(\textbf{\textit{V}}_{X}^{H})+\text { upsample }(f^{L \rightarrow H}(\textbf{\textit{V}}_{X}^{L}), 2) \\
\textbf{\textit{V}}_{Y}^{L}=&\textbf{\textit{V}}_{Y}^{L \rightarrow L}+\textbf{\textit{V}}_{Y}^{H \rightarrow L} \\
=&f^{L \rightarrow L}(\textbf{\textit{V}}_{X}^{L} )+f^{H \rightarrow L}(\operatorname{pool}(\textbf{\textit{V}}_{X}^{H}, 2) )
\end{aligned},
\end{equation}
where $ H \rightarrow H $ and $ L \rightarrow L $ denote the intra-frequency information update, $ L \rightarrow H $ and $ H \rightarrow L $ denote the inter-frequency communication, $ f(\textbf{\textit{V}}_{X} )$ denotes a  convolution, $ upsample (f(\textbf{\textit{V}}_{X}), k)$ is an up-sampling operation by a factor of $ k $ via nearest interpolation, $ pool (\textbf{\textit{V}}_{X}, k)$  is an average pooling operation with kernel size $k \times k$ and stride $ k $.

Compared to the regular convolution operation, the OctConv can learn the difference of low- and high-frequency between fake and real faces more effectively. Moreover, OctConv can capture more frequency information by enlarging the receptive field size due to reduce the spatial resolution of the low-frequency feature maps. Based on the above analysis, the OctConv is suitable for learning features from CDI and SI related to frequency domain information, thus we combined the OctConv and the successful CNN network ResNet-34 to mine intrinsic features effectively.

%\begin{equation}
%\begin{array}{l}
%V_{Y}^{H}=f^{H \rightarrow H}\left(V_{X}^{H}\right)+f^{L \rightarrow H}\left(V_{X}^{L}\right) \\
%\left.V_{Y}^{L}=f^{L \rightarrow L}\left(V_{X}^{L}\right)+f^{H \rightarrow L}\left(V_{X}^{H}\right)
%\end{array}, 
%\end{equation}

\begin{figure*}[ht]
	\centering
	\includegraphics[scale=0.35]{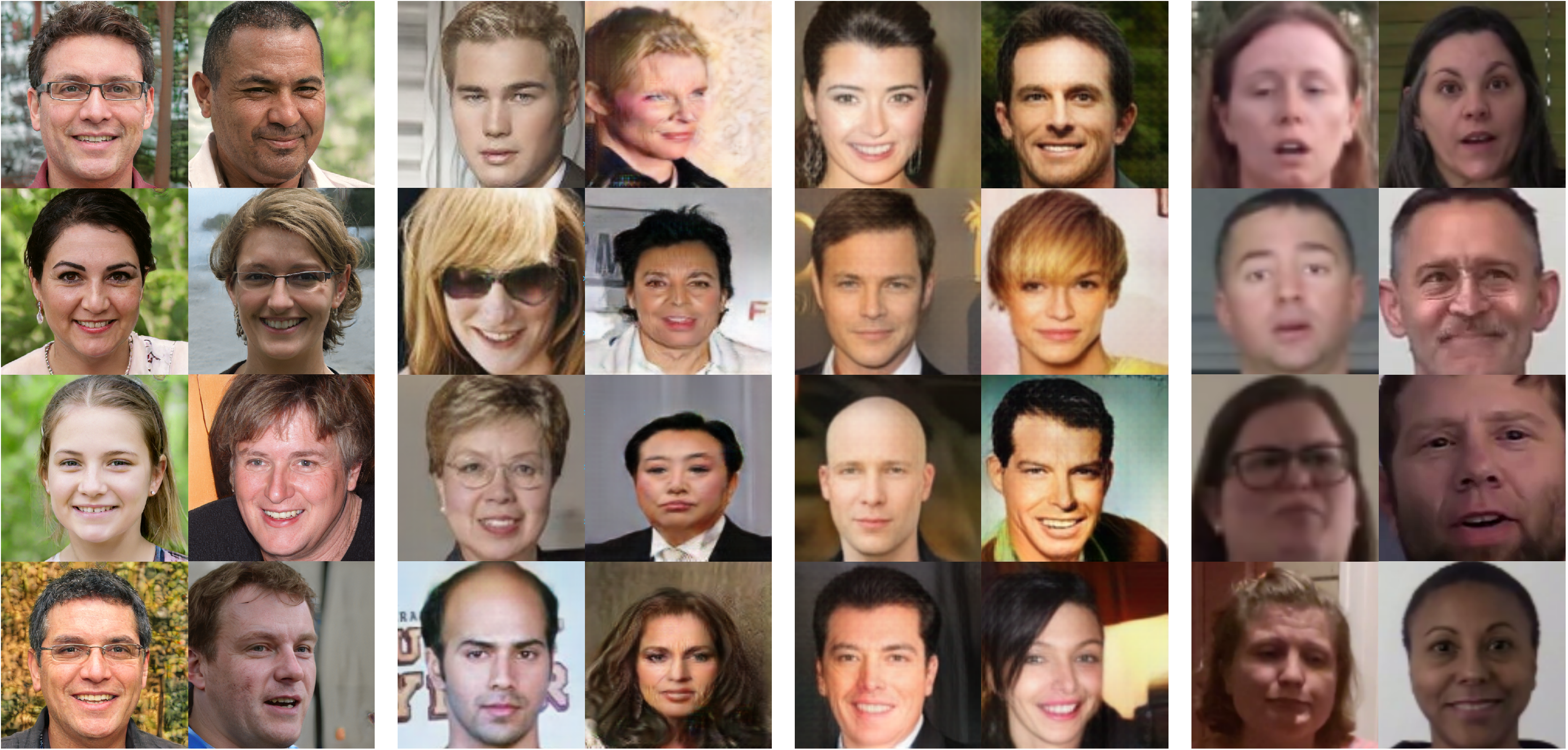}
	\caption{Examples of AI-manipulated fake faces.}
	\label{example_dataset}
\end{figure*}

\subsubsection{ Attention-based Feature Fusion}
After learning two stream information, we can obtain the features $\textbf{\textit{V}}_{CDI}$ and $\textbf{\textit{V}}_{SI}$. We introduce a dot-product attention-based fusion module to effectively fuse these two features, rather than exploiting the simple fusion methods (e.g., feature concatenation) that cannot deeply explore the interplay of features from different sources. Specifically, as shown in Fig.~\ref{framework} (b), we first learn a kernel $ q $ which has the same dimension of $\textbf{\textit{V}}_{CDI}$ and $\textbf{\textit{V}}_{SI}$. $ q $ is used to filter the features via dot product. And then the filter generates vectors $\textbf{\textit{D}}_{CDI}$ and $\textbf{\textit{D}}_{SI}$ that represent the significance of the corresponding feature could be expressed as:
\begin{equation}
\begin{array}{l}
\textbf{\textit{D}}_{CDI}=q^{T} \textbf{\textit{V}}_{CDI}\\
\textbf{\textit{D}}_{SI}=q^{T} \textbf{\textit{V}}_{SI}
\end{array}.
\end{equation}
Next, the $\textbf{\textit{D}}_{CDI}$ and $\textbf{\textit{D}}_{SI}$ are passed to a softmax operator to obtain the corresponding weights $W_{CDI}$ and $W_{SI}$. Finally, the fusion feature can be formulated as follows:
\begin{equation}
\textbf{\textit{V}}_{F}=\textbf{\textit{V}}_{CDI}\otimes W_{CDI} \oplus \textbf{\textit{V}}_{SI}\otimes W_{SI}, 
\end{equation}
where $ \otimes $ is element-wise multiplication, and $\oplus $ is concatenation for these two features.

Different from simple fusion methods, the proposed dot-product attention fusion module can adaptively fuse these two features, thereby helping our framework focus on useful information autonomously to mine intrinsic features.

\begin{table*}[]
	\centering
	\caption{ Details of our Four Categories of Dataset.}
	\label{table1}
	\renewcommand\arraystretch{1.9}
	\begin{tabular}{|c|c|c|c|c|c|c|}
		\hline
		\textbf{Dataset}                               & \textbf{Manipulation Techniques} & \textbf{Collection}                            & \textbf{Total Samples} & \textbf{Real Faces}              & \textbf{Total Samples} & \textbf{Size}                         \\ \hline
		\multirow{2}{*}{Entire Synthesis}     & StyleGAN             & \multirow{2}{*}{Officially- released} & 5000          & \multirow{2}{*}{FFHQ}   & 5000          & \multirow{2}{*}{1024 $ \times $ 1024} \\ \cline{2-2} \cline{4-4} \cline{6-6}
		& StyleGAN2            &                                       & 5000          &                         & 5000          &                              \\ \hline
		\multirow{3}{*}{Expression Manipulation} & ExperGAN             & \multirow{3}{*}{Self-synthesis}       & 5000          & \multirow{3}{*}{CelebA} & 5000          & \multirow{3}{*}{128 $ \times $ 128}   \\ \cline{2-2} \cline{4-4} \cline{6-6}
		& GANimation           &           & 5000          &                         & 5000          &                              \\ \cline{2-2} \cline{4-4} \cline{6-6}
		& HomoInterpGAN        &                                       & 5000          &                         & 5000          &                              \\ \hline
		\multirow{3}{*}{Attribute Manipulation}  & CycleGAN             & \multirow{3}{*}{Self-synthesis}       & 5000          & \multirow{3}{*}{CelebA} & 5000          & \multirow{3}{*}{128 $ \times $ 128}   \\ \cline{2-2} \cline{4-4} \cline{6-6}
		& StarGAN              &                                       & 5000          &                         & 5000          &                              \\ \cline{2-2} \cline{4-4} \cline{6-6}
		& STGAN                &                                       & 5000          &                         & 5000          &                              \\ \hline
		\multirow{2}{*}{Identity Manipulation}        & Faceswap             & FaceForensics++                       & 15000         & FaceForensics++         & 15000         & \multirow{2}{*}{256 $ \times $ 256}   \\ \cline{2-6}
		& DeepFake             & DFDC                                  & 30000         & DFDC                    & 30000         &                              \\ \hline
	\end{tabular}
\end{table*}

\subsection{Domain Alignment Module}
\label{generalization}
In detecting AI-manipulated fake faces, although existing detection methods consider different categories of fake faces, the detection performance on unseen manipulation techniques is still poor. The main reason is that on the one hand, the intrinsic features of fake faces are not well captured, on the other hand, there are differences among cross-manipulation techniques in feature distribution. Therefore, besides the previous module for mining intrinsic features, we also need to eliminate the distribution bias. In view of this, we propose a domain alignment module by minimizing the divergence among cross-manipulation techniques in feature distribution to reduce the distribution bias. The basic idea is that we enforce the domain alignment module to learn generalized features to further improve the generalization ability, as shown in Fig.~\ref{framework} (c).

Based on the analysis above, we first partition faces into different domains according to different manipulation techniques. Specifically, the face samples with $ K $ seen domains for training are denoted as:
\begin{equation}
\textit{\textbf{X}}_{d}=\left[\textbf{\textit{x}}_{d_{1}}, \ldots, \textbf{\textit{x}}_{d_{n_{d}}}\right],
\end{equation} 
where domain $d \in\{1, \ldots, K\}$ and $ n_{d} $ is the number of samples of the domain $ d $. The corresponding labels are denoted as: $\textbf{\textit{Y}}_{d}=\left[\textbf{\textit{y}}_{d_{1}}, \ldots, \textbf{\textit{y}}_{d_{n_{d}}}\right]$ with 2 categories (fake/real). After mining intrinsic features, the input features of the $ m $-th fully-connected layer is described as: 
\begin{equation}
\hat{\textbf{\textit{V}}}_{F}^{m}=\left[\textbf{\textit{V}}_{F_{1}}^{m}, \ldots, \textbf{\textit{V}}_{F_{K}}^{m}\right],
\end{equation}
where $ \textbf{\textit{V}}_{F_{d}}^{m} \in \mathbb{R}^{n_{d} \times D}$ denotes the input features of the $ m $ -th fully connected layer from domain $ d $. To eliminate the bias among cross-manipulation techniques in feature distribution, we propose a cross-domain alignment loss:
	\begin{equation}
	 \mathcal{L}_{\mathrm{CDA}}= d\left(\hat{\textbf{\textit{V}}}_{F}^{m}\right), 
	\end{equation}
where $d()$ indicates the distance among cross-manipulation techniques. We use the Maximum Mean Discrepancy (MMD) \cite{gretton2012kernel} measure in our work:
\begin{equation}
d\left(\hat{\textbf{\textit{V}}}_{F}^{m}\right)=
\frac{1}{K(K-1)} \sum_{d \neq j}\left\|\frac{1}{n_{d}} \sum_{t_{1}=1}^{n_{d}} \textbf{\textit{V}}_{F_{d},t_{1}}^{m}-\frac{1}{n_{j}} \sum_{t_{2}=1}^{n_{j}} \textbf{\textit{V}}_{F_{j}, t_{2}}^{m}\right\|^{2}
\end{equation}
where $\textbf{\textit{V}}_{F_{d},t}^{m}\in \mathbb{R}^{D}$ denotes the input feature of $ t $ -th sample from $\textbf{\textit{V}}_{F_{d}}^{m}$. Therefore, we train our framework from scratch on the face samples collected from multiple domains with cross-entropy loss $\mathcal{L}_{\mathrm{C}}$ \cite{de2005tutorial} and cross-domain alignment loss $\mathcal{L}_{\mathrm{CDA}}$. In the cross-domain alignment loss $\mathcal{L}_{\mathrm{CDA}}$, the MMD distance among the domains is required to be minimized. In other words, the network parameters can be learned as:
\begin{equation}
\Theta^{*}=\arg \min _{\Theta}\   \mathcal{L}_{\mathrm{C}}+\lambda \mathcal{L}_{\mathrm{CDA}}, 
\end{equation}
where $\lambda$ is the weight of the cross-domain alignment loss.

Through explicitly restricting the bias among cross-manipulation techniques in feature distribution, our proposed framework is guided to learn domain-invariant representations to further improve generalization ability for detecting fake faces with unseen manipulation techniques.
	
%$X_{i}=\left[{X}_{1}, {X}_{2}, \ldots, {X}_{K}\right]$, corresponding labels are denoted as $Y=\left[{Y}_{1}, {Y}_{2}, \ldots, {Y}_{K}\right]$ with 2 categories (fake/real), the total number of samples in $ X $ is $N_{1}+N_{2}+\ldots+N_{K}$, where $ N_{i} $ represent the number of samples from the domain $ i $. 

%Therefore, we train our framework from scratch on the face samples collected from multiple domains with cross-entropy loss $\mathcal{L}_{\mathrm{C}}$ \cite{de2005tutorial} and cross-domain alignment loss $\mathcal{L}_{\mathrm{CDA}}$. In the domain alignment loss $\mathcal{L}_{\mathrm{a}}$, the MMD distance among the domains is required to be minimized. As such, the network parameters can be learned as:
%\begin{equation}
%\Theta^{*}=\arg \min _{\Theta}\   \mathcal{L}_{\mathrm{c}}+\lambda \mathcal{L}_{\mathrm{a}}, 
%\end{equation}
%where $ \mathcal{L}_{\mathrm{a}}= d\left(\bm{\hat{F}_{f, CDSI}}\right) $, $\lambda$ is the weight of the domain alignment loss. The gradient of the domain alignment loss $\mathcal{L}_{\mathrm{a}}$ with respect to the network parameter $\Theta$ can be computed as:
%\begin{equation}
%\frac{\partial d\left(\bm{\hat{F}_{f, CDSI}}\right)}{\partial \Theta}=\frac{\partial d\left(\bm{\hat{F}_{f, CDSI}}\right)}{\partial \bm{\hat{F}_{f, CDSI}}} \frac{\partial \bm{\hat{F}_{f, CDSI}}}{\partial \Theta}, 
%\end{equation}
%where $\frac{\partial \bm{\hat{F}_{CDSI}}}{\partial \Theta}$ can be obtained by back-propagation method.

\begin{table*}[]
	\centering
	\caption{Performance on The Detection of Fake Faces with Seen Manipulation Techniques in Four Categories of Dataset (\%).}
	\label{table2}
	\renewcommand\arraystretch{1.7}
	\begin{tabular}{|c|c|c|c|c|}
		\hline
		\textbf{Dataset}                     & \textbf{Face Entire Synthesis}& 	\textbf{Facial Expression Manipulation} & 	\textbf{Facial Attribute Manipulation}&\textbf{Identity Manipulation}\\ \hline
		Mesonet \cite{afchar2018mesonet}      & 89.93   & 90.95    & 91.45   & 89.99     \\ \hline
		Capsule \cite{nguyen2019capsule}  & 91.78   & 93.32    & 92.21   & 90.06      \\ \hline
		Xception \cite{rossler2019faceforensics++}   & 93.05   & 92.97    & 92.05   & 91.08      \\ \hline
		Color Spaces \cite{he2019detection}        & 93.23   & 93.3    & 93.28   & 91.38       \\ \hline
		Saturation Cues \cite{mccloskey2019detecting}     & 94.2   & 93.42    & 92.9   & 91.34       \\ \hline
		Gram-Net \cite{liu2020global}               & 94.05   & 95.12    & 94.1   & 92.84       \\ \hline
		Self-Attention \cite{mi2020gan}           & 95.7   & 95.55    & 96.53   & 94.65       \\ \hline
		FakeSpotter \cite{wang2019fakespotter}       & 96.65   & 96.88    & 97.47   & 95.68       \\ \hline
		$ F^{3} $-Net \cite{qian2020thinking}       & 96.83   & 97.13    & 97.77   & 97.42       \\ \hline
		\textbf{Proposed}                                              & \textbf{98.95} & \textbf{99.57} & \textbf{99.75} & \textbf{98.97}  \\ \hline
	\end{tabular}
\end{table*}

\begin{table*}[]
	\centering
	\caption{Performance on The Detection of Fake Faces with Seen Data in Each Manipulation Techniques (\%).}
	\label{table8}
	\renewcommand\arraystretch{1.9}
	\setlength{\tabcolsep}{0.37mm}{
		\begin{tabular}{|c|c|c|c|c|c|c|c|c|c|c|}
			\hline
			\textbf{Manipulation Techniques}                                & \textbf{StyleGAN}       & \textbf{StyleGAN2 }     & \textbf{ExperGAN}       & \textbf{GANimation}     & \textbf{HomoInterpGAN} & \textbf{CycleGAN  }     & \textbf{StarGAN }       & \textbf{STGAN  }        &\textbf{ Faceswap }      &\textbf{ DeepFake  }     \\ \hline
			Mesonet \cite{afchar2018mesonet}              & 90.2           & 89.66          & 90.4           & 90.65          & 91.8          & 90.55          & 91.15          & 92.65          & 90.02          & 89.98          \\ \hline
			Capsule \cite{nguyen2019capsule}              & 92.1           & 91.46          & 92.9           & 93.15          & 93.9          & 92.65          & 91.93          & 92.05          & 90.13          & 90.03          \\ \hline
			Xception \cite{rossler2019faceforensics++}    & 93.25          & 92.85          & 93.1           & 92.85          & 92.95         & 92.9           & 91.8           & 91.45          & 91.15          & 91.04          \\ \hline
			Color Spaces \cite{he2019detection}           & 93.55          & 92.9           & 92.95          & 93.1           & 93.85         & 93.95          & 92.85          & 93.05           & 91.77          & 91.18          \\ \hline
			Saturation Cues \cite{mccloskey2019detecting} & 94.95          & 93.45          & 93.8           & 93.25          & 93.2          & 92.95          & 92.9           & 92.85          & 91.48          & 91.27          \\ \hline
			Gram-Net \cite{liu2020global}                 & 94.8           & 93.3           & 95.15          & 95.25          & 94.95         & 94.7           & 93.75          & 93.85          & 93.25          & 92.63          \\ \hline
			Self-Attention \cite{mi2020gan}                & 96.1           & 95.3           & 96.15          & 95.4           & 95.1          & 96.55          & 96.3           & 96.75          & 94.25          & 94.85          \\ \hline
			FakeSpotter \cite{wang2019fakespotter}        & 96.9           & 96.4           & 96.5           & 96.75          & 97.4          & 97.15          & 97.3           & 97.95          & 96.72          & 95.16          \\ \hline
			$ F^{3} $-Net \cite{qian2020thinking}                & 97.1           & 96.55           & 96.75          & 97.2           & 97.45          & 97.3          & 97.9           & 98.1          & 98.08          & 97.09          \\ \hline
			\textbf{Proposed}                                              & \textbf{99.25} & \textbf{98.65} & \textbf{99.45} & \textbf{99.55} & \textbf{99.7} & \textbf{99.87} & \textbf{99.73} & \textbf{99.65} & \textbf{99.67} & \textbf{98.62} \\ \hline
	\end{tabular}}
\end{table*}

\section{Experimental Results}
\label{experiment}

\subsection{Datasets}
The dataset used for experiments is constructed by four categories: face entire synthesis, facial expression manipulation, facial attribute manipulation and identity manipulation, and our dataset contains 10 state-of-the-art and popular manipulation techniques. For each dataset, we collect fake faces by generating them with the manipulation techniques or downloading the officially
released generated faces. To make the distribution of the real and fake faces as
close as possible, real faces are pre-processed according
to the pipeline prescribed by each technique. Specifically, for the face entire synthesis dataset (with the size 1024 $\times $ 1024), 10000 real face images are collected from FFHQ \cite{karras2019style}, and 5000 synthesis faces are collected from two publicy datasets with StyleGAN and StyleGAN2, respectively. For the facial expression manipulation dataset (with the size 128 $\times $ 128), 15000 real faces are collected from CelebA \cite{A1}, and corresponding 5000 manipulation faces generated by ExperGAN, GANimation and HomoInterpGAN, respectively. For the facial attribute manipulation dataset (with the size 128 $\times $ 128), 15000 real faces are collected from CelebA, and corresponding 5000 manipulation faces generated by CycleGAN, StarGAN and STGAN, respectively. For the identity manipulation dataset (with the size 256 $\times $ 256), we utilize the FaceSwap dataset of Faceforensics++ \cite{rossler2019faceforensics++} and DeepFake dataset of DFDC \footnotemark[2]. Specifically, we first collect 1500 FaceSwap videos with different compression coefficient c0, c23 and c40, 3000 DeepFake videos and the equal number of real videos. After that, we extract 10 frames from each video, and further use face detector MTCNN \cite{zhang2016joint} to get the images of face region. If there are more than one face detected in a frame, only the largest one is extracted, and then the extracted faces are aligned to size of 256 $\times $ 256. Consequently, 15000 manipulation face images with FaceSwap, 30000 manipulation face images with DeepFake and the equal number of real face images are produced. Details of our dataset are summarized in Table~\ref{table1} and examples of AI-manipulated fake faces are shown in Fig.~\ref{example_dataset}.
\footnotetext[2]{https://www.kaggle.com/c/deepfake-detection-challenge} 

% \usepackage{multirow}

% \usepackage{multirow}

\begin{table*}[]
	\centering
	\caption{Details of Cross-Manipulation Technique Experimental Protocols.}
	\label{table9}
	\renewcommand\arraystretch{1.1}
	\setlength{\tabcolsep}{1.4mm}{
		\begin{tabular}{|c|c|c|c|c|c|}
			\hline
			\textbf{Protocol} & \textbf{\begin{tabular}[c]{@{}c@{}}Training Manipulated \\ Techniques\end{tabular}}    & \textbf{\begin{tabular}[c]{@{}c@{}}Testing Manipulated \\ Techniques\end{tabular}} & \textbf{Protocol} & \textbf{\begin{tabular}[c]{@{}c@{}}Training Manipulated \\ Techniques\end{tabular}} & \textbf{\begin{tabular}[c]{@{}c@{}}Testing Manipulated \\ Techniques\end{tabular}} \\ \hline
			N1                & \begin{tabular}[c]{@{}c@{}}StyleGAN\\ ExperGAN\\ CycleGAN\\ Faceswap\\GANimation\end{tabular}       & StyleGAN2                                                                           & N5                & \begin{tabular}[c]{@{}c@{}}StyleGAN2\\ GANimation\\ STGAN\\DeepFake\end{tabular}            & StyleGAN                                                                            \\ \hline
			N2                & \begin{tabular}[c]{@{}c@{}}GANimation\\ StarGAN\\ DeepFake\\ StyleGAN\\Faceswap\end{tabular} & ExperGAN                                                                            & N6                & \begin{tabular}[c]{@{}c@{}}StyleGAN2\\HomoInterpGAN\\ CycleGAN\\ Faceswap\\\end{tabular}           & GANimation                                                                          \\ \hline
			N3                & \begin{tabular}[c]{@{}c@{}}StyleGAN2\\ CycleGAN\\ ExperGAN\\ DeepFake\\STGAN\end{tabular}      & STGAN                                                                               & N7                & \begin{tabular}[c]{@{}c@{}}DeepFake\\ExperGAN\\ STGAN\\ StyleGAN\end{tabular}             & StarGAN                                                                             \\ \hline
			N4                & \begin{tabular}[c]{@{}c@{}}Faceswap\\ StyleGAN2\\ STGAN\\ HomoInterpGAN\\CycleGAN\end{tabular}          & DeepFake                                                                            & N8                & \begin{tabular}[c]{@{}c@{}}StyleGAN\\ HomoInterpGAN\\ DeepFake\\CycleGAN\end{tabular}          & Faceswap                                                                            \\ \hline
	\end{tabular}}
\end{table*}

\begin{table*}[]
	\centering
	\caption{Performance on The Detection of Fake Faces with Unseen Manipulation Techniques in Four Categories of Datase (\%).}
	\label{table3}
	\renewcommand\arraystretch{2.4}
	\setlength{\tabcolsep}{3.5mm}{
		\begin{tabular}{|c|c|c|c|c|c|c|c|c|}
			\hline
			\textbf{Protocol} & \textbf{N1}    & \textbf{N2}    & \textbf{N3}    & \textbf{N4}    & \textbf{N5}    & \textbf{N6}    & \textbf{N7}    & \textbf{N8}    \\ \hline
			Zhang \textit{et al}. \cite{zhang2019detecting}             & 90.82          & 91.39          & 91.17         & 90.74          & 90.19         & 90.72          & 90.57          & 90.07        \\ \hline
			Li \textit{et al}. \cite{li2020identification}                 & 92.09         & 92.63          & 92.79          & 91.93          & 91.15          & 92.08          & 91.92          & 90.76          \\ \hline
			Dang \textit{et al}. \cite{dang2020detection}              & 93.59          & 94.82          & 95.29          & 93.45          & 92.62          & 93.21          & 94.04          & 92.51          \\ \hline
			Wang \textit{et al}. \cite{wang2020cnn}               & 95.78          & 96.18          & 96.24          & 95.34         & 94.87          & 96.15          & 96.01          & 94.67          \\ \hline
			Chai \textit{et al}. \cite{chai2020makes}               & 97.05          & 97.92          & 97. 76         & 96.29          & 96.63          & 97.18          & 96.79          & 96.52          \\ \hline
			\textbf{Proposed} & \textbf{97.92} & \textbf{98.92} & \textbf{98.59} & \textbf{97.65} & \textbf{97.26} & \textbf{98.13} & \textbf{98.09} & \textbf{97.19} \\ \hline
	\end{tabular}}
\end{table*}

\subsection{Implementation Details}
The GPU card utilized for our task is NVIDIA GTX2080Ti and the framework is implemented by Torch library. The input size is 128 $\times $ 128, and we crop each image to several nonoverlapping 128 $\times $ 128 images. During training, we first train the framework only with cross-entropy loss, and then the last convolutional layer as well as the fully connected layer are fine-tuned with both cross-entropy loss and cross-domain alignment loss. Such training strategy is reasonable since shallow layers are more likely to be generalized \cite{yosinski2014transferable} and more discriminative information expected to extract by fine-tuning the deeper layers. For the learning parameter setting, we use the Adam optimizer \cite{kingma2014adam} in a mini-batch manner with the size 10 during network initialization step and 100 for each domain during the fine-tuning step. The momentum values are set as $ \beta_{1} $ = 0.9 and $ \beta_{2} $= 0.999, and the initial learning rate is set to $1 \mathrm{e}-3$ and $1 \mathrm{e}-4$ for fine-tuning. In addition, learning rates are dropped by 10 $\times $ if after 5 epochs the validation accuracy does not increase by 0.1\%, and we terminate training at learning rate $1 \mathrm{e}-7$. The weight $\lambda$ of the cross-domain alignment loss is set in the way that at the end of training, the classification loss and cross-domain alignment loss are approximately the same. The reason for such setting is that our framework can learn both the discrimination and generalization ability. More specifically, the weight $\lambda$ is selected in a range $ \{0.001, 0.01, 0.1, 1, 10\} $. During testing, a query face is first cropped to several 128 $\times $ 128 parts, from which we can recover the entire face image. Then, these cropped 128 $\times $ 128 parts will be judged by independently. Finally, the face image is considered to be fake if one of the cropped parts is judged to be fake. In our experiments, 5 experiments are performed and the average result is reported each time. 

\subsection{Detection of Fake Faces with Seen Manipulation Techniques}
In this section, we evaluate the performance of our framework on the detection of fake faces with seen manipulation techniques in all four datasets, that is, the manipulation technique in the testing sets also exists in the training sets. We merge four datasets together and the training, validation and testing sets are randomly selected. The ratio of face images in training, validation and testing sets is set to be 6:2:2 for each dataset. In order to better demonstrate the superiority of the proposed method, we compare the results with the state-of-the-art methods on detection of AI-manipulated fake faces. For comparison, their methods are applied to our datasets. The comparative results on four categories of fake faces are shown in Table~\ref{table2}. From the results, all the accuracies of our framework are higher than 98\%, and outperform state-of-the-art methods in all cases. We further evaluate the performance of our framework on each manipulation techniques and compare the results with state-of-the-art methods, as shown in Table~\ref{table8}. Experimental results demonstrate the excellent performance of our method, as well as a significant improvement compared with the state-of-the-art methods. Especially on the latest DFDC and StyleGAN2 datasets, we also achieve relatively high accuracies. This could be explained as below: compared to these methods, our framework not only exploits intrinsic clues from SI and CDI based on the fundamental differences between the AI-manipulation process and the camera imaging process, but also introduces the OctConv-based feature learning module and the attention-based feature fusion module to effectively mine intrinsic features for fake faces detection.

In summary, the proposed framework can effectively detect the AI-manipulated fake faces with seen manipulation techniques.

\subsection{Detection of Fake Faces with Unseen Manipulation Techniques}

In this section, to further evaluate the generalization capability of our proposed framework, we conduct experiments to evaluate the performance on the detection of fake faces with unseen manipulation techniques. Specifically, we first train a framework by employing fake faces produced by multiple manipulation techniques and corresponding real faces. Then we evaluate the performance by testing faces with another manipulation technique which is not involved in the training phase. To conduct such unseen-manipulation technique based experiments, we merge four datasets together and rearrange training, validation and testing sets based on manipulation techniques. In particular, we randomly create 8 different cross-manipulation technique scenarios to evaluate the performance of our framework, and the details of experimental protocols are demonstrated in Table~\ref{table9}. Moreover, we compare the result with state-of-the-art methods focusing on generalization to demonstrate the generalization ability of our framework. For comparison, their methods are applied to our dataset. The comparative results are shown in Table~\ref{table3}. It is observed that although the accuracies of detecting unseen manipulation techniques are slightly decreased than that of detecting seen manipulation techiniques, but the results are all over 97\%. Moreover, it can be obtained from the results that our proposed framework outperforms state-of-the-art methods in all cases, which proves that our framework has better generalization ability, and the generalization performance is better with more source domains: (Protocol N1-N4). This can be explained by the following: the universal detector or simulator may not be able to simulate the artifacts of all manipulation techniques in \cite{zhang2019detecting} and \cite{wang2020cnn}. Moreover, compared to \cite{dang2020detection}, \cite{li2020identification} and \cite{chai2020makes}, we mine more efficient intrinsic features and propose an alignment module to obtain generalized features for fake faces detection.

In a word, the proposed framework shows excellent advantages in improving generalization performance.

\begin{table}[]
	\centering
	\caption{Details of The Dataset for Ablation Study.}
	\label{table4}
	\renewcommand\arraystretch{1.8}
	\setlength{\tabcolsep}{1.4mm}{
		% \usepackage{multirow}
		\begin{tabular}{|c|c|c|c|c|}
			\hline
			&              & \textbf{Training Set} & \textbf{Validation Set} & \textbf{Test Set}\\ \hline
			\multirow{2}{*}{Real Faces} & CelebA        & 3000         & 1000           & 0        \\ \cline{2-5} 
			& FFHQ          & 3000         & 1000           & 0        \\ \hline
			\multirow{5}{*}{Fake Faces} & StyleGAN      & 3000         & 1000           & 0        \\ \cline{2-5} 
			& StyleGAN2     & 0            & 0              & 2000     \\ \cline{2-5} 
			& ExperGAN      & 3000         & 1000           & 0        \\ \cline{2-5} 
			& GANimation    & 3000         & 1000           & 0        \\ \cline{2-5} 
			& HomoInterpGAN & 0            & 0              & 2000     \\ \hline
	\end{tabular}}
\end{table}

\begin{table}[]
	\centering
	\caption{The Results on RGB Feature, CDI Feature, SI Feature, Simple Concatenation Feature, and Attention Fusion Feature (\%).}
	\label{table6}
	\renewcommand\arraystretch{1.8}
	\begin{tabular}{|c|c|c|c|c|c|}
		\hline
		& $\textbf{\textit{V}}_{RGB}$& $\textbf{\textit{V}}_{CDI}$ & $\textbf{\textit{V}}_{SI}$ & $\textbf{\textit{V}}_{C}$& $\textbf{\textit{V}}_{F}$  \\ \hline
		StyleGAN2     & 91.85 & 92.55             & 93.75           & 95.60              & \textbf{98.75}              \\ \hline
		HomoInterpGAN  & 92.15  & 94.20              & 93.25           & 96.85              &  \textbf{99.65}         \\ \hline
	\end{tabular}
\end{table}

\subsection{Ablation Study}
In this section, we perform the ablation study to evaluate the performance of each component of our framework. We use the face entire synthesis and facial expression synthesis datasets, and the specific details of the dataset is shown as Table~\ref{table4}. 

\subsubsection{Results on Different Features }
As described in Section ~\ref{CDSI}, our framework learn intrinsic features from CDI and SI instead of RGB images. To verify that the CDI and SI contain more intrinsic clues, we exploit the feature $\textbf{\textit{V}}_{RGB}$ learned from RGB images to detect fake faces on the dataset for ablation study. Moreover, the features $\textbf{\textit{V}}_{CDI}$ and $\textbf{\textit{V}}_{SI}$ are combined into $ \textbf{\textit{V}}_{F} $ with the attention fusion module. Therefore, in order to analyze these two feature sets, we utilize $\textbf{\textit{V}}_{CDI}$ and $\textbf{\textit{V}}_{SI}$ to detect fake faces on the dataset used for ablation study. We further compare the simple concatenation fusion feature $\textbf{\textit{V}}_{C}$ and the attention fusion feature $ \textbf{\textit{V}}_{F} $ to evaluate the performance of the attention fusion module. The results are shown in Table~\ref{table6}. From the results, the detection results based on $\textbf{\textit{V}}_{CDI}$ and $\textbf{\textit{V}}_{SI}$ achieve better performances than that on $\textbf{\textit{V}}_{RGB}$, due to the clues of CDI and SI are efficient and intrinsic. It is also observed that the results on the simple concatenation fusion feature $\textbf{\textit{V}}_{C}$ are all better than those on the feature set $\textbf{\textit{V}}_{CDI}$ or $\textbf{\textit{V}}_{SI}$. Moreover, the results on the attention fusion feature $ \textbf{\textit{V}}_{F} $ are all best compared to those on other features, which demonstrate the necessity of attention feature fusion module.

\begin{figure}[ht]
	\centering
	\includegraphics[scale=0.68]{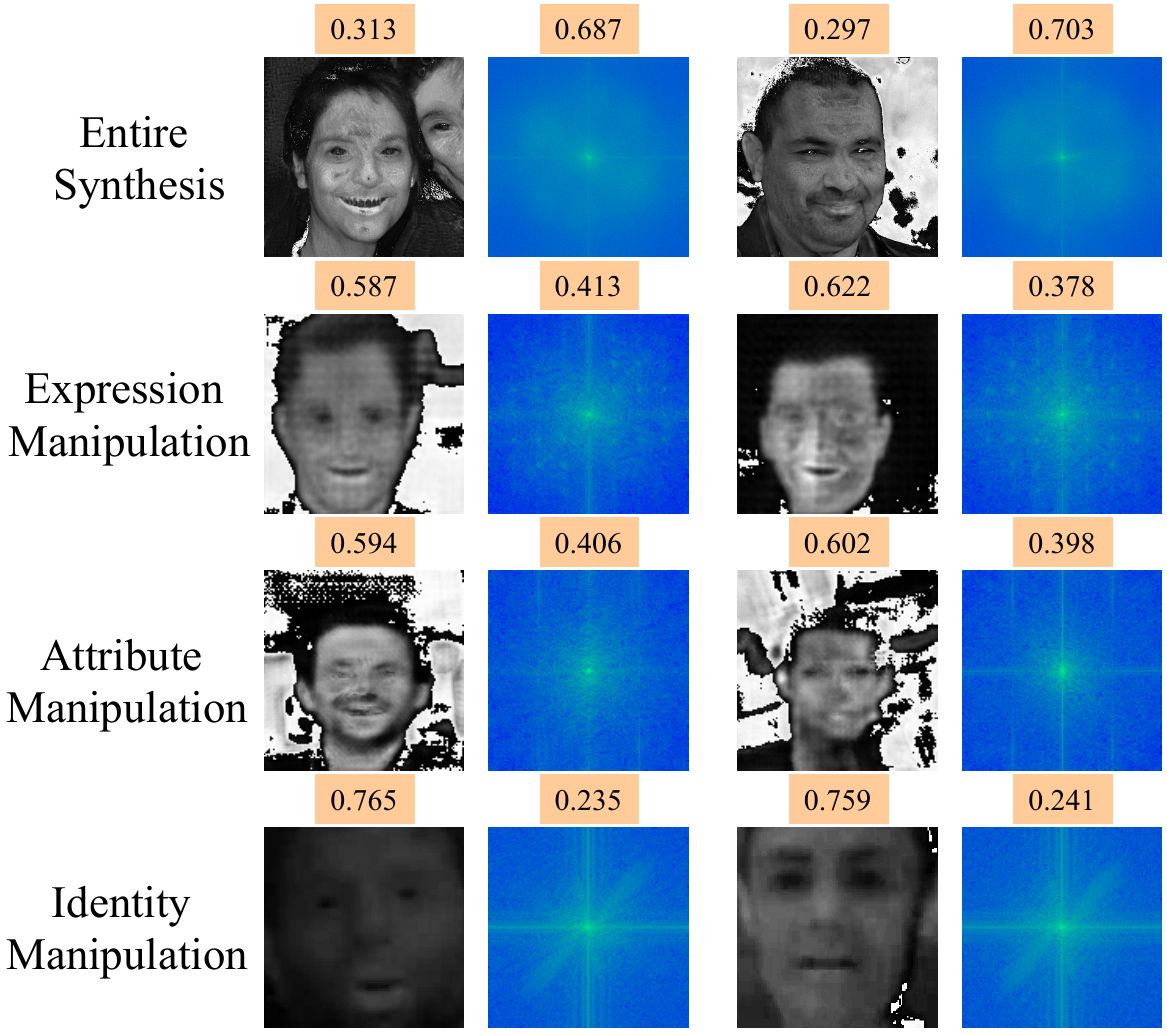}
	\caption{Attention fusion weights (numbers in the boxes) showing the importance of CDI and SI. Samples
		cover four categories of fake faces.}
	\label{avisual}
\end{figure}

\subsubsection{ Visualizations of Attention-based Feature Fusion Module}
To explore the effectiveness of our attention-based feature fusion module, we further show the visualization results. Some samples are selected from four categories of fake faces datasets for analyzing the adaptive weighting mechanism of the fusion module. From the samples in Fig.~\ref{avisual}, we can see the weights for CDI and SI are adaptively weighting. For the entire synthesis faces, the weights of SI are higher than those of CDI because the clues from SI is more efficient. For the expression manipulation and attribute manipulation faces, the weights for CDI and SI are similar, and the weights of CDI are slightly higher than those of SI. For the identity manipulation faces, the CDI gain higher weights because of more pre- and post-processing in the video forgery process.

\subsubsection{Importance of OctConv Operator}
In order to evaluate the performance of the octconv operator, we test the framework with and without octconv operator on the dataset used for ablation study. The comparison results are shown in Table~\ref{table5}. From the Table~\ref{table5}, the framework with OctConv operator performs better, which prove the effectiveness of the OctConv Operator.

\begin{table}[]
	\centering
	\caption{ The Results of Ablation Study on Octconv Operator (\%).}
	\label{table5}
	\renewcommand\arraystretch{1.8}
	\begin{tabular}{|c|c|c|}
		\hline
		& Without OctConv & \textbf{With OctConv} \\ \hline
		StyleGAN2     & 96.45              & \textbf{98.75}          \\ \hline
		HomoInterpGAN & 96.75              & \textbf{99.65}           \\ \hline
	\end{tabular}
\end{table}

\subsubsection{Importance and Visualizations of Domain Alignment Module}
In this section, we further verify the effectiveness of domain alignment module in our framework. We test with and without the domain alignment module on the dataset used for ablation study. The comparison results are shown in Table~\ref{table7}. From the results, the domain alignment module increase the performance of the proposed framework, and it plays a key role in improving the generalization ability.

\begin{table}[]
	\centering
	\caption{ The Results of Ablation Study on Domain Alignment Module (DAM)(\%).}
	\label{table7}
	\renewcommand\arraystretch{1.8}
	\setlength{\tabcolsep}{1.5mm}{
		\begin{tabular}{|c|c|c|}
			\hline
			& Without DAM & \textbf{With DAM}\\ \hline
			StyleGAN2     & 92.10                                                                     & \textbf{98.75}                                                                  \\ \hline
			HomoInterpGAN & 94.35                                                                     & \textbf{99.65}                                                     \\ \hline
	\end{tabular}}
\end{table}

\begin{figure}[ht]
	\centering
	\includegraphics[scale=0.365]{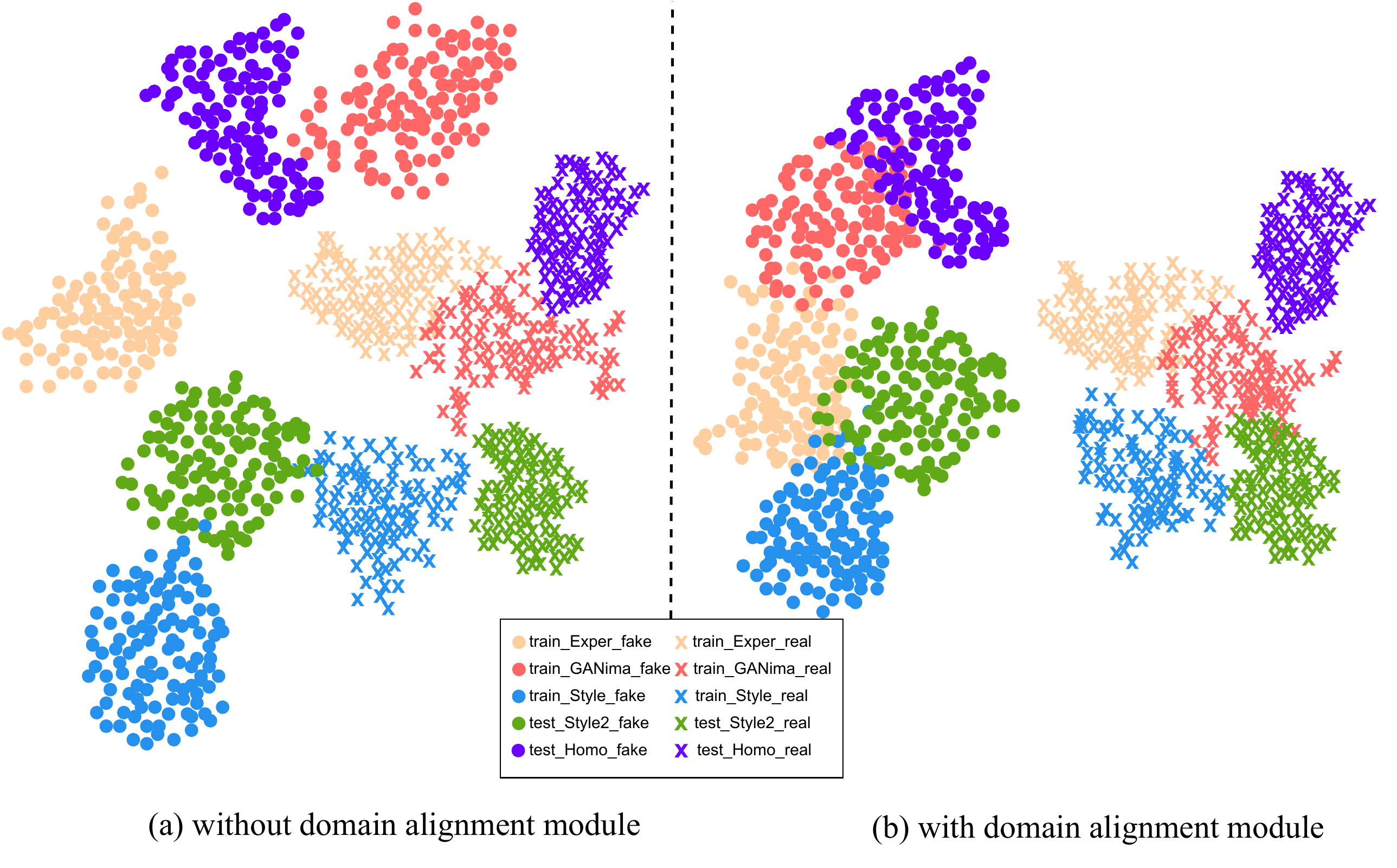}
	\caption{The t-SNE visualizations of the features from the dataset for ablation study (best viewed in color). (a) Features visualization without the domain alignment module. (b) Features visualization with the proposed domain alignment module.}
	\label{domainvisual}
\end{figure}

Moreover, we randomly select 200 samples of each manipulation technique from the dataset for ablation study and plot the t-SNE \cite{maaten2008visualizing} visualizations to analyze the feature space learned by introducing the domain alignment module, as shown in Fig.~\ref{domainvisual}. It can be seen that the domain alignment module can make the features more robust in the feature space, which is more conducive to detect fake faces with unseen manipulation techniques.

\section{Conclusion And Future Work}
\label{con}
In this paper, we propose a novel framework to effectively detect AI-manipulated fake faces, especially focus on how to mine the generalized features on detecting unseen manipulation techniques.
To achieve the goal, we mine intrinsic features and further eliminate the distribution bias among cross-manipulation techniques. First, we mine two intrinsic clues from the CDI and SI, rather than depending on the specific defects in the manipulation process. Moreover, we adopt OctConv and an attention-based fusion module to mine two intrinsic features more effectively. Finally, to obtain a more generalized framework, an alignment module is proposed to reduce the bias among cross-manipulation techniques in feature distribution. The experimental results show that the performance of the proposed method outperforms the state-of-the-art works, especially the generalization performance on detecting unseen manipulation techniques. In the future work, we will consider more challenging situations and focus on the effectiveness in detecting AI-manipulated fake faces on social networks.

\ifCLASSOPTIONcaptionsoff
  \newpage
\fi

\bibliographystyle{IEEEtran} 
\bibliography{re}
%\begin{IEEEbiography}[{\includegraphics[width=1in,height=1.25in,clip,keepaspectratio]{yy}}]{Michael Shell}
\begin{IEEEbiography}[{\includegraphics[width=1in,height=1.25in,clip,keepaspectratio]{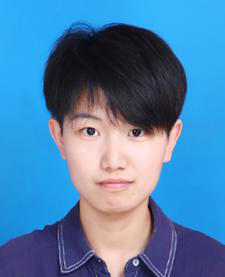}}]{Yang Yu}
	received the B.S. degree from China
University of Petroleum, Beijing, China, in 2017.
She is pursuing the Ph.D. degree in Beijing Jiaotong University, China.
\end{IEEEbiography}

\begin{IEEEbiography}[{\includegraphics[width=1in,height=1.25in,clip,keepaspectratio]{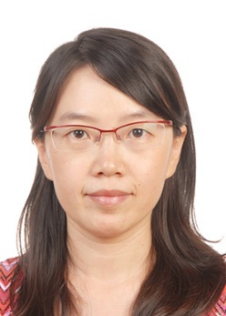}}]{Rongrong Ni}
	received the Ph.D. degree in signal and information processing from the Institute of Information Science, Beijing Jiaotong University, Beijing, China, in 2005. Since 2005, she has been the faculty of the School of Computer and Information Technology and the Institute of Information Science, where she was an Associate Professor from 2008 to 2013, and was promoted to a Professor in 2013.
	
	Her current research interests include image processing, data hiding and digital forensics, pattern recognition, and computer vision.
\end{IEEEbiography}
\begin{IEEEbiography}[{\includegraphics[width=1in,height=1.25in,clip,keepaspectratio]{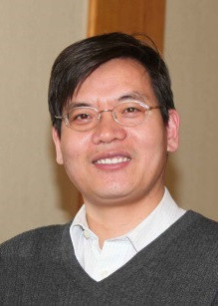}}]{Yao Zhao}
	received the B.S. degree from Fuzhou University, Fuzhou, China, in 1989, and the M.E. degree from Southeast University, Nanjing, China, in 1992, both from the Radio Engineering Depart- ment, and the Ph.D. degree from the Institute of Information Science, Beijing Jiaotong University (BJTU), Beijing, China, in 1996. He became an Associate Professor at BJTU in 1998 and became a Professor in 2001. From 2001 to 2002, he was a Senior Research Fellow with the Information and Communication Theory Group, Faculty of Information Technology and Systems, Delft University of Technology, Delft, The Netherlands.
	He is currently the Director of the Institute of Information Science, BJTU. His current research interests include image/video coding, digital watermarking and forensics, and video analysis and understanding.
\end{IEEEbiography}

%\newpage

\end{document}